\documentclass[11pt]{article}

\usepackage{acl2023}

\usepackage{times}
\usepackage{latexsym}
\usepackage{enumitem}

\usepackage{stfloats}

\usepackage[T1]{fontenc}

\usepackage[utf8]{inputenc}

\usepackage{microtype}

\usepackage[tight,footnotesize]{subfigure}

\usepackage{graphicx} 
\graphicspath{{Figures/}}

\usepackage{booktabs}
\usepackage{multirow}
\usepackage{bm}
\usepackage{amssymb}
\usepackage{amsmath}
\usepackage{xspace}

\usepackage{ctable}
\usepackage{csquotes}

\usepackage{graphicx} 
\newcommand{\ie}{\emph{i.e.,}\xspace}
\newcommand{\eg}{\emph{e.g.,}\xspace}

\newcommand{\paratitle}[1]{\vspace{0.8ex}\noindent \textbf{#1}}

\usepackage{pifont}

\usepackage{stmaryrd}

\usepackage{soul}

\title{Rethinking Document-Level Relation Extraction:  A Reality Check}

\author{Jing Li$^1$, Yequan Wang$^2$, Shuai Zhang$^3$ and Min Zhang$^{1*}$\\
  $^1$Harbin Institute of Technology, Shenzhen, China \\
  $^2$Beijing Academy of Artificial Intelligence, Beijing, China\\
  $^3$ETH Zurich, Switzerland \\
  \texttt{\{li.jing, zhangmin2021\}@hit.edu.cn,}\\ \texttt{tshwangyequan@gmail.com,  cheungdaven@gmail.com} \\}

\begin{document}
\maketitle
\begin{abstract}
	
Recently, numerous efforts have continued to push up performance boundaries of document-level relation extraction (DocRE) and have claimed significant progress in DocRE. 
In this paper, we do not aim at proposing a novel model for DocRE. 
Instead, we take a closer look at the field to see if these performance gains are actually true. 
By taking a comprehensive literature review and a thorough examination of popular DocRE datasets, we find that these performance gains are achieved upon a strong or even untenable assumption in common: all named entities are perfectly localized, normalized, and typed in advance. 
Next, we construct four types of entity mention attacks to examine the robustness of typical DocRE models by behavioral probing.
We also have a close check on model usability in a more realistic setting. 
Our findings reveal that most of current DocRE models are vulnerable to entity mention attacks and difficult to be deployed in real-world end-user NLP applications. 
Our study calls more attentions for future research to stop simplifying problem setups, and to model DocRE in the wild rather than in an unrealistic Utopian world.

\let\thefootnote\relax\footnotetext{$^*$ Corresponding author.}
\end{abstract}

\begin{figure}[t]		
	\centering
	\includegraphics[width=0.95\columnwidth,draft=false]{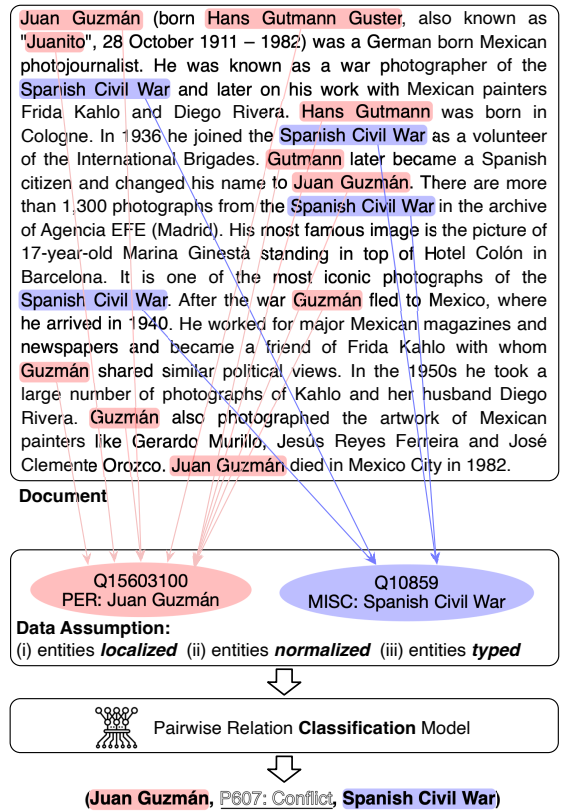}
	\caption{Data assumption in most of DocRE models.   }
	\label{fig:dataexample}
\end{figure}
\section{Introduction}
Document-level relation extraction (DocRE), aiming at identifying semantic relations between a head entity and a tail entity in a document~\cite{DBLP:conf/acl/YaoYLHLLLHZS19}, plays an essential role in a variety of downstream applications,  such as question answering~\cite{DBLP:conf/acl/XuRFHZ16} and knowledge base construction~\cite{DBLP:conf/acl/TrisedyaWQZ19}.

Recently, there are two flourishing branches for DocRE. 
First, graph-based approaches consider entities~\cite{DBLP:conf/iclr/VelickovicCCRLB18,DBLP:conf/acl/NanGSL20}, mentions~\cite{DBLP:conf/emnlp/ChristopoulouMA19,DBLP:conf/coling/LiYSXXZ20} and sentences \cite{DBLP:conf/aaai/XuCZ21} as nodes to construct a document-level graph and perform reasoning through some advanced neural graph techniques. 
Second, sequence-based approaches leverage BiLSTM \cite{DBLP:conf/acl/HuangZF0L020,DBLP:conf/acl/LiXLFRJ21} or Transformers \cite{DBLP:conf/acl/TanHBN22,DBLP:conf/naacl/ZhongC21,DBLP:journals/corr/abs-2102-01373} as encoders to learn document-level representations.  
However, all these models have one thing in common that they are based on \textbf{a strong or even untenable assumption} as shown in Figure~\ref{fig:dataexample}: all entity mentions are (i) correctly localized; (ii) perfectly normalized; (iii)  correctly typed.\footnote{More illustrating examples can be found in Appx. \S \ref{app:moreillustrating}}
Then, the task of modeling DocRE is usually simplified as a pairwise \textbf{classification} problem.

Although these pairwise classification approaches have claimed significant progress in DocRE performance, we are still interested in taking a closer look at the field to see if this is actually true.
In particular, many research papers have reported very decent leaderboard scores for the DocRE task.
Does this mean the task of DocRE has been almost completely solved?
Can the current approaches be widely used in real-world DocRE scenarios?

To answer these questions, we first take a closer look at data annotations in commonly-used DocRE datasets to check the strong data assumption (\S \ref{sec:dataannotation}).
We focus specifically on the annotations of named entity recognition (NER) and normalization (\ie entity linking) in detecting relations. 
By answering three research questions (\texttt{\textbf{RQ1-3}}), we find that current problem setups for DocRE are greatly simplified and unrealistic. 

If the data assumption is too strict, it is not clear whether current DocRE models are robust in a variety of loose assumptions. 
Therefore, we construct four types of attacks regarding entity mention annotations to investigate the model robustness (\S \ref{sec:robustness}, \texttt{\textbf{RQ4}}) using behavioral probing~\cite{DBLP:conf/acl/LasriPLPC22, DBLP:conf/naacl/0003HNC22}. 

To further have a look at the limitations of data assumptions, it is important to investigate the usability of existing DocRE models in real-world scenarios. 
Hence, we examine the capability of widely-used NER systems and entity linking systems on preparing model input formats from raw text for DocRE model deployment (\S \ref{sec:usability}, \texttt{\textbf{RQ5}}). 
Finally, we discuss our empirical findings and call special attentions for future research in developing DocRE models (\S \ref{sec:discussions}). 

In short, our contributions and findings are: 
\begin{itemize}[noitemsep,nolistsep]
	\item We present a comprehensive literature review on recent advances for DocRE and identify a strong or even untenable assumption in modeling DocRE.  
	\item We take a thorough examination of data annotation on three popular DocRE datasets.  Detecting relations in text commonly involves multiple mentions and aliases of paired entities (\ie head and tail entities) which are currently assumed to be perfectly typed, localized and normalized before modeling DocRE. 
	\item We construct four types of entity mention attacks to check the robustness for typical DocRE models. Most of current DocRE models are vulnerable to mention attacks (F1 drops from 7.93\% to 85.51\%). 
	\item We have a close check on the usability of typical DocRE models. Under the identified data assumption, current DocRE models are very difficult to be deployed in real-world end-user NLP applications because of the need of input preparation for each pipeline module (\ie the reproduction rate of input format is only from 34.3\% to 58.1\%). 
	\item We discuss our findings, and call attentions for future research to stop simplifying problem setups,  and to model DocRE in the wild rather than in an unrealistic Utopian world. 
\end{itemize}

\begin{table*}[t]
	\centering
	\small
	\scalebox{0.96}{
		\begin{tabular}{@{}cccccccc@{}}
			\toprule
			\multirow{2}{*}{\textbf{References}} &	\multirow{2}{*}{\textbf{Venue}} & \multirow{2}{*}{\textbf{Claim}} & \multirow{2}{*}{\textbf{Performed}} & \multicolumn{3}{c}{\textbf{Annotation Assumption}}   &\multirow{2}{*}{\textbf{Aggregation}}             \\ \cmidrule(l){5-7} 
			&                        &                   &    & \textbf{Localization} & \textbf{Linking} & \textbf{Typing} \\ \cmidrule(r){1-8}
					\cite{DBLP:journals/corr/abs-2211-14470}	& EMNLP22 &  Extraction                     &    Classification                   &     \ding{51}         &          	\ding{51}   & 	\ding{55}        & LogSumExp         \\
			\cite{DBLP:conf/acl/XieSLM022}	& ACL22 &  Extraction                     &    Classification                   &     \ding{51}         &          	\ding{51}   & 	\ding{55}        & LogSumExp         \\
			
			\cite{DBLP:conf/acl/TanHBN22}	&  ACL22                      &   Extraction                    &     Classification           &   	\ding{51}    &   	\ding{51}           & 	\ding{55}        & LogSumExp          \\
			
			\cite{DBLP:journals/corr/abs-2109-12093}	& NAACL22 &  Extraction                     &    Classification                   &     \ding{51}         &          	\ding{51}   & 	\ding{51}        & LogSumExp         \\
			\cite{DBLP:journals/corr/abs-2204-12679}	&  NAACL22                      &   Extraction                    &     Classification           &   	\ding{51}    &   	\ding{51}           & 	\ding{51}        & Average          \\
			
			\cite{yurelation}	&  NAACL22                      &   Extraction                    &     Classification           &   	\ding{51}    &   	\ding{51}           & 	\ding{55}        & Average          \\

			\cite{DBLP:conf/acl/ZengWC21}&         ACL21               &   Extraction                    &    Classification            &  	\ding{51}    &	\ding{51}    &	\ding{51}               &        Average       \\
			
			\cite{DBLP:conf/acl/LiXLFRJ21}	&       ACL21                  &   Extraction                    &     Classification         &   	\ding{51}    &    	\ding{51}                  &    	\ding{51}              &     Max-pooling                     \\

			\cite{DBLP:conf/acl/XuCZ21}&      ACL21  &   Extraction                    &     Classification                &    \ding{51}                     &    \ding{51}              &     \ding{51}          &      Average         \\

			\cite{DBLP:conf/acl/HuangZF0L020}&  ACL21                 &   Extraction                    &     Classification                &    \ding{51}                     &    \ding{51}              &     \ding{55}   &       Average\\
			
			\cite{DBLP:conf/acl/MakinoMS21}&  ACL21                 &   Extraction                    &     Classification                &    \ding{51}                     &    \ding{51}              &     \ding{51}   &       Max-pooling\\

			\cite{DBLP:conf/emnlp/RuSFQ000021} &EMNLP21                 &   Extraction                    &     Classification                &    \ding{51}                     &    \ding{51}              &     \ding{51}   &       Average\\

			\cite{DBLP:conf/emnlp/ZhangWUJNP21} &EMNLP21                 &   Extraction                    &     Classification                &    \ding{51}                     &    \ding{51}              &     \ding{51}   &       [CLS]\\

			\cite{DBLP:conf/ijcai/ZhangCXDTCHSC21}&IJCAI21                 &   Extraction                    &     Classification                &    \ding{51}                     &    \ding{51}              &     \ding{55}   &       LogSumExp\\
			
			\cite{DBLP:conf/aaai/XuCZ21}&AAAI21                 &   Extraction                    &     Classification                &    \ding{51}                     &    \ding{51}              &     \ding{55}   &       Average\\
			
			\cite{xu2021entity}&AAAI21                 &   Extraction                    &     Classification                &    \ding{51}                     &    \ding{51}              &     \ding{51}   &       Average\\

			\cite{DBLP:conf/aaai/LiYHZ21}&AAAI21                 &   Extraction                    &     Classification                &    \ding{51}                     &    \ding{51}              &     \ding{51}   &       Average\\
			
			\cite{DBLP:conf/aaai/Zhou0M021}&AAAI21                 &   Extraction                    &     Classification                &    \ding{51}                     &    \ding{51}              &     \ding{55}   &       Average\\

			\cite{DBLP:conf/acl/NanGSL20}&ACL20                 &   Extraction                    &     Classification                &    \ding{51}                     &    \ding{51}              &     \ding{55}   &       Average\\
			
			\cite{DBLP:conf/emnlp/ZengXCL20}&EMNLP20                 &   Extraction                    &     Classification                &    \ding{51}                     &    \ding{51}              &     \ding{51}   &       Average\\
			
			\cite{DBLP:conf/emnlp/WangHCS20}&EMNLP20                 &   Extraction                    &     Classification                &    \ding{51}                     &    \ding{51}              &     \ding{51}   &       Average\\
			
			\cite{DBLP:conf/emnlp/TranNN20}&EMNLP20                 &   Extraction                    &     Classification                &    \ding{51}                     &    \ding{51}              &     \ding{51}   &       Average\\

			\cite{DBLP:conf/coling/LiYSXXZ20}&COLING20                 &   Extraction                    &     Classification                &    \ding{51}                     &    \ding{51}              &     \ding{51}   &       Average\\
			
			\cite{DBLP:conf/coling/ZhangYSLTWG20}&COLING20                 &   Extraction                    &     Classification                &    \ding{51}                     &    \ding{51}              &     \ding{51}   &       Average\\
			
			\cite{DBLP:conf/coling/ZhouXYLLJ20}&COLING20                 &   Extraction                    &     Classification                &    \ding{51}                     &    \ding{51}              &     \ding{51}   &       Average\\
			
			\cite{DBLP:conf/emnlp/ChristopoulouMA19}&EMNLP19                 &   Extraction                    &     Classification                &    \ding{51}                     &    \ding{51}              &     \ding{51}   &       Average\\	
			
			\cite{DBLP:conf/naacl/JiaWP19}&NAACL19                 &   Extraction                    &     Classification                &    \ding{51}                     &    \ding{51}              &     \ding{55}   &       LogSumExp\\	
			
			\bottomrule
	\end{tabular}}
	\caption{Recent DocRE models in anti-chronological order. ``\textbf{Localization}'', ``\textbf{Linking}'' and ``\textbf{Typing}'' indicates that an approach needs accurate annotations of entity localization, entity linking and entity typing, respectively. ``\textbf{Aggregation}'' indicates the strategy that how to aggregate  multiple mention representations of an entity.}
	\label{tab:relatedsurvey}
\end{table*}

\section{A Quick Literature Review}
In this section, we have a quick literature review of DocRE models to shed light on a global review for recent evolutions. 
Table~\ref{tab:relatedsurvey} summarizes recent studies in anti-chronological order.

\paratitle{Graph-based Approaches.} 
Graph-based approaches first construct a  document-level homogeneous graph where words \cite{DBLP:conf/coling/ZhangYSLTWG20}, mentions  \cite{DBLP:conf/emnlp/ChristopoulouMA19}, entities  \cite{DBLP:conf/coling/ZhouXYLLJ20}, sentences \cite{DBLP:conf/coling/LiYSXXZ20,xu2021entity} or meta dependency paths \cite{DBLP:conf/acl/NanGSL20} are considered as nodes and some semantic dependencies (\eg mention-mention \cite{DBLP:conf/emnlp/ChristopoulouMA19},  mention-entity \cite{DBLP:conf/emnlp/ZengXCL20}, mention-sentence \cite{DBLP:conf/emnlp/WangHCS20}, entity-sentence \cite{DBLP:conf/coling/LiYSXXZ20}, sentence-sentence \cite{DBLP:conf/emnlp/WangHCS20,DBLP:conf/acl/XuCZ21}, sentence-document \cite{DBLP:conf/acl/ZengWC21}) as edges. One key advantage of these approaches is that some advanced graph techniques can be used to model inter- and intra-entity interactions and perform multi-hop reasoning.

\paratitle{Sequence-based Approaches.} Instead of introducing complex graph structures, some 
approaches typically model a document as a sequence of tokens and leverage BiLSTM \cite{DBLP:conf/acl/HuangZF0L020,DBLP:conf/acl/LiXLFRJ21} or Transformers \cite{DBLP:conf/acl/TanHBN22} as encoder to capture the contextual semantics. 
In particular, some studies have already contributed effort to integrating entity structures \cite{DBLP:conf/aaai/XuCZ21}, concept view \cite{DBLP:conf/aaai/LiYHZ21}, deep probabilistic logic \cite{DBLP:conf/emnlp/ZhangWUJNP21}, U-shaped Network \cite{DBLP:conf/ijcai/ZhangCXDTCHSC21}, relation-specific attentions \cite{yurelation},  logic rules \cite{DBLP:conf/emnlp/RuSFQ000021}, augmenting intermediate steps \cite{DBLP:journals/corr/abs-2109-12093}, sentences importance estimation \cite{DBLP:journals/corr/abs-2204-12679}, evidence extraction \cite{DBLP:conf/acl/XieSLM022} and knowledge distillation \cite{DBLP:conf/acl/TanHBN22}
 into transformer-based neural models. 
In addition, some studies \cite{DBLP:conf/acl/SoaresFLK19,DBLP:conf/aaai/Zhou0M021,DBLP:conf/naacl/ZhongC21,DBLP:journals/corr/abs-2102-01373,DBLP:journals/corr/abs-2211-14470} already verified that inserting special symbols (\eg $[\text{entity}]$ and $[\text{/entity}]$) before and after named entities can significantly benefit relation representation encoding.

\paratitle{Observations from Literature Review.} From Table ~\ref{tab:relatedsurvey}, we have following key observations:  (1) The listed studies claim that they address the problem of  ``document-level relation \textbf{extraction}''\footnote{The term ``Extraction'' commonly refers to extract relation types, head and tail entities from raw text.}, but the relation \textbf{classification} is actually performed. 
(2) All graph-based approaches build homogeneous or heterogeneous graphs based on the unrealistic \textbf{precondition} that accurate annotations of entity localization, entity linking and entity typing are available. 
(3) Some pooling strategies (\eg Max, Average and LogSumExp) are widely used in modeling DocRE when aggregating representations of multiple mentions of an entity.  However, it is unclear how the wrongly-detected mentions affect the performance of DocRE models.

\section{Check on Dataset Annotations}
\label{sec:dataannotation}

To provide in-depth observations of the data assumption in most of DocRE models, 
we first take a thorough examination of data annotations on three commonly-used DocRE datasets.
We will conduct quantitative and qualitative studies to analyze entity mentions and entity aliases which a relation instance involved.\footnote{Entity Mentions: The words in text that refer to an entity. Entity Aliases:  Unique mentions of an entity.  Relation Instance: A piece of text involving head and tail entities to be classified. }

\subsection{Probing Datasets}

\begin{figure*}[t]
	\centering
	\subfigure[DocRED] {\includegraphics[width=0.32\textwidth,draft=false]{ 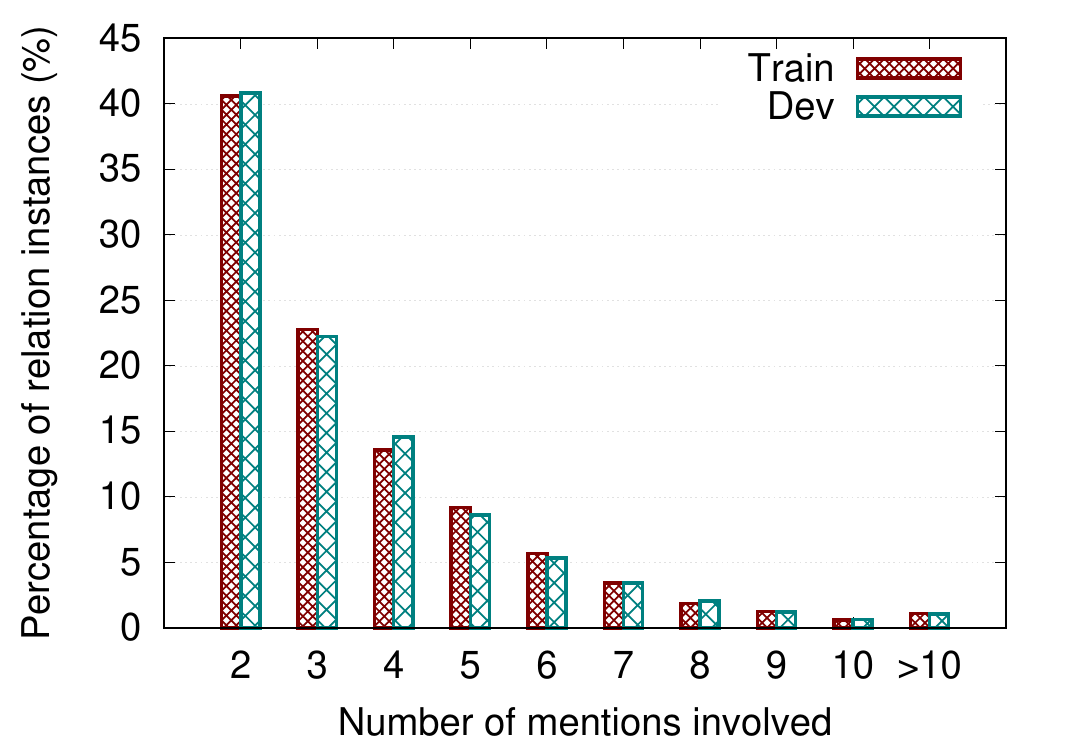}\label{subfig:docred}} 
	\subfigure[CDR] {\includegraphics[width=0.32\textwidth,draft=false]{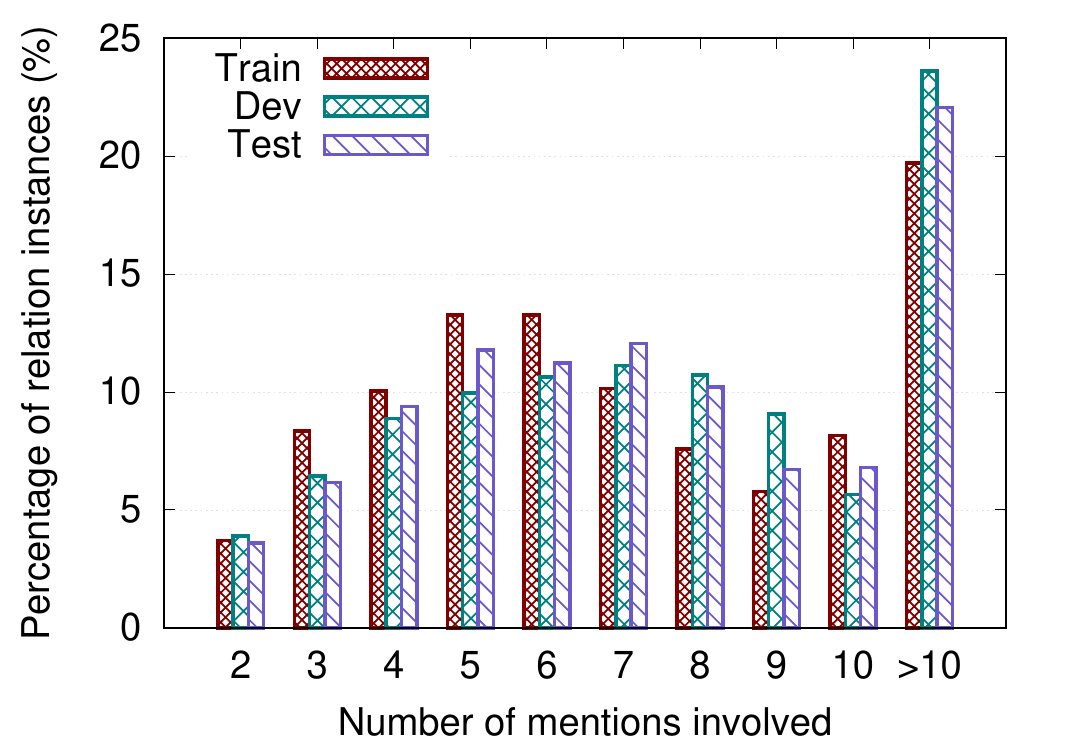} \label{subfig:cdr}}  
	\subfigure[GDA] 	{\includegraphics[width=0.32\textwidth, draft=false]{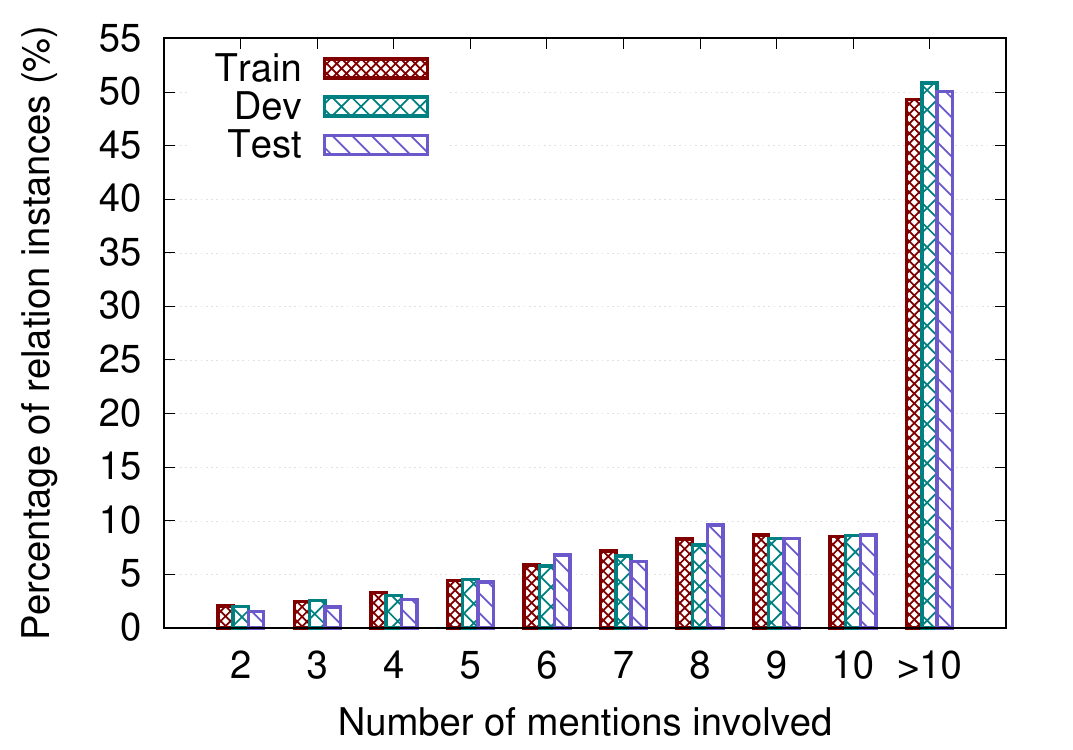}\label{subfig:gda}} 
	\caption{Entity mention statistics on three datasets.}
	\label{fig:percent}
\end{figure*}

\begin{table}
	\centering
	\small
	\scalebox{1}{
		\begin{tabular}{c|r|r|r|r}
			\toprule
			\multicolumn{2}{c|}{\textbf{Datasets}} & \textbf{\#Doc.} & \textbf{\#Rel.} & \textbf{\#Non-NA}\\ \midrule
			\multirow{3}{*}{\textbf{DocRED}} & Train & 3,053  & 97  & 385,272    \\
			& Dev & 1,000  &   97&   11,518\\
			& Test &1,000   &  97 &  - \\ \hline
			\multirow{3}{*}{\textbf{CDR}} & Train &  500 & 2  &  1,055   \\
			&Dev & 500  &2   &1,025    \\
			& Test &500   & 2  & 1,087   \\ \hline
			\multirow{3}{*}{\textbf{GDA}} & Train &23,353   & 2  &  36,079  \\
			& Dev &  5,839 &2   &8,762  \\
			& Test&  1,000 & 2  & 1,502  \\ \bottomrule
	\end{tabular}}
	\caption{Statistics of DocRE datasets. (\#Doc.: number of documents, \#Rel.: number of relation labels,	\#Non-NA: number of non-NA-relation instances.)}
	\label{tab:datasets}
\end{table}

The summary of datasets is shown in Table~\ref{tab:datasets}. 
NA-instance means that there is no relation between head and tail entities. 
Non-NA instance means that there is at least one relation between head and tail entities. 
Note that the mention statistics in this Section are based on Non-NA instances.

\paratitle{DocRED}~\cite{DBLP:conf/acl/YaoYLHLLLHZS19} is a human-annotated dataset from Wikipedia and Wikidata.  
DocRED has 5,053 documents, 97 relation classes, 132,275 entities, and 56,354 relational facts in total.
The average length of documents in DocRED is around 8 sentences. 
Following previous studies~\cite{DBLP:conf/acl/YaoYLHLLLHZS19,DBLP:journals/corr/abs-1909-11898}, we use the standard split of the dataset: 3,053 documents for training, 1,000 for
development and 1,000 for test.

\paratitle{CDR}~\cite{DBLP:journals/biodb/LiSJSWLDMWL16} consists of three separate sets of articles with diseases, chemicals and their relations annotated. There are two relation labels: None and Chemical-Disease. 
There are total 1,500 articles and 500 each for the training, development and test sets.

\paratitle{GDA}~\cite{DBLP:conf/recomb/WuLLTL19} is a Gene-Disease Association dataset from MEDLINE abstracts: 29,192 articles for training and 1,000 for testing. Following previous studies~\cite{DBLP:conf/emnlp/ChristopoulouMA19,DBLP:conf/acl/LiXLFRJ21}, we further split the original training set into two sets:  23,353 for training and 5,839 for development. 
There are two relation labels: None and Gene-Disease.

\subsection{Data Observations and Findings}
We organize our findings by answering following research questions (RQs):

\paratitle{\texttt{\textbf{(RQ1)}}}: \textbf{How many entity mentions are involved in a relation instance in commonly-used DocRE datasets?}

We define that a relation instance to be classified is a piece of text containing head and tail entities. 
Thus, it is natural that the head or tail entity may have multiple  mentions in the document. 
Figure~\ref{subfig:docred}, \ref{subfig:cdr} and \ref{subfig:gda} show entity mention statistics in DocRED, CDR and GDA, respectively. 
The horizontal axis shows number of mentions of a relation instance. 
The vertical axis shows the percentages of relation instances in datasets. 
In the DocRED dataset, 59.2\% of relation instances have more than two mentions. 
For CDR,  96\% of relation instances have more than two mentions and 21\% of relation instances have more than 10 mentions. 
For GDA, 98\% of relation instances have more than two mentions and 50\% of relation instances have more than 10 mentions. 
Our this finding reveals the huge difference between the sentence-level and document-level RE. 
That is, document-level RE involves much more entity mentions than sentence-level RE because of the longer text in document-level RE. 
One strong (almost untenable) assumption of existing DocRE models is that all entity mentions of a relation instance are successfully identified.

\paratitle{\texttt{\textbf{(RQ2)}}}: \textbf{How many aliases does an entity have in commonly-used DocRE datasets?}

\texttt{\textbf{RQ1}} already showed that a relation instance may have multiple entity mentions. 
A follow-up question is about the number of unique mentions. 
Given that an entity can appear multiple times in a document, we define that the aliases of an entity are unique mentions. 
We are interesting in how many aliases an entity has. 

\begin{figure*}[t]
	\centering
	\subfigure[DocRED] { \includegraphics[width=0.32\textwidth,draft=false]{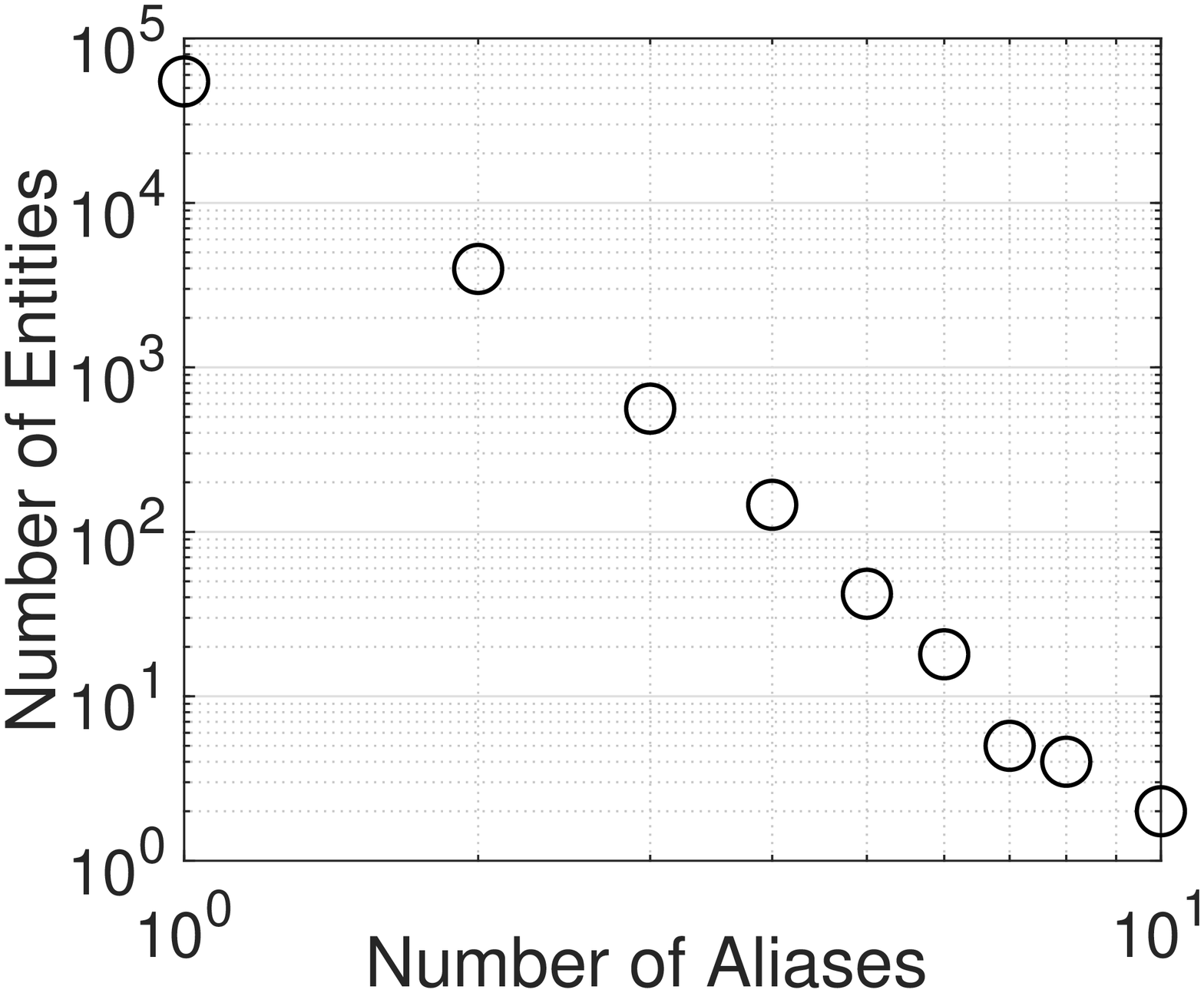} }  
	\subfigure[CDR] { \includegraphics[width=0.32\textwidth,draft=false]{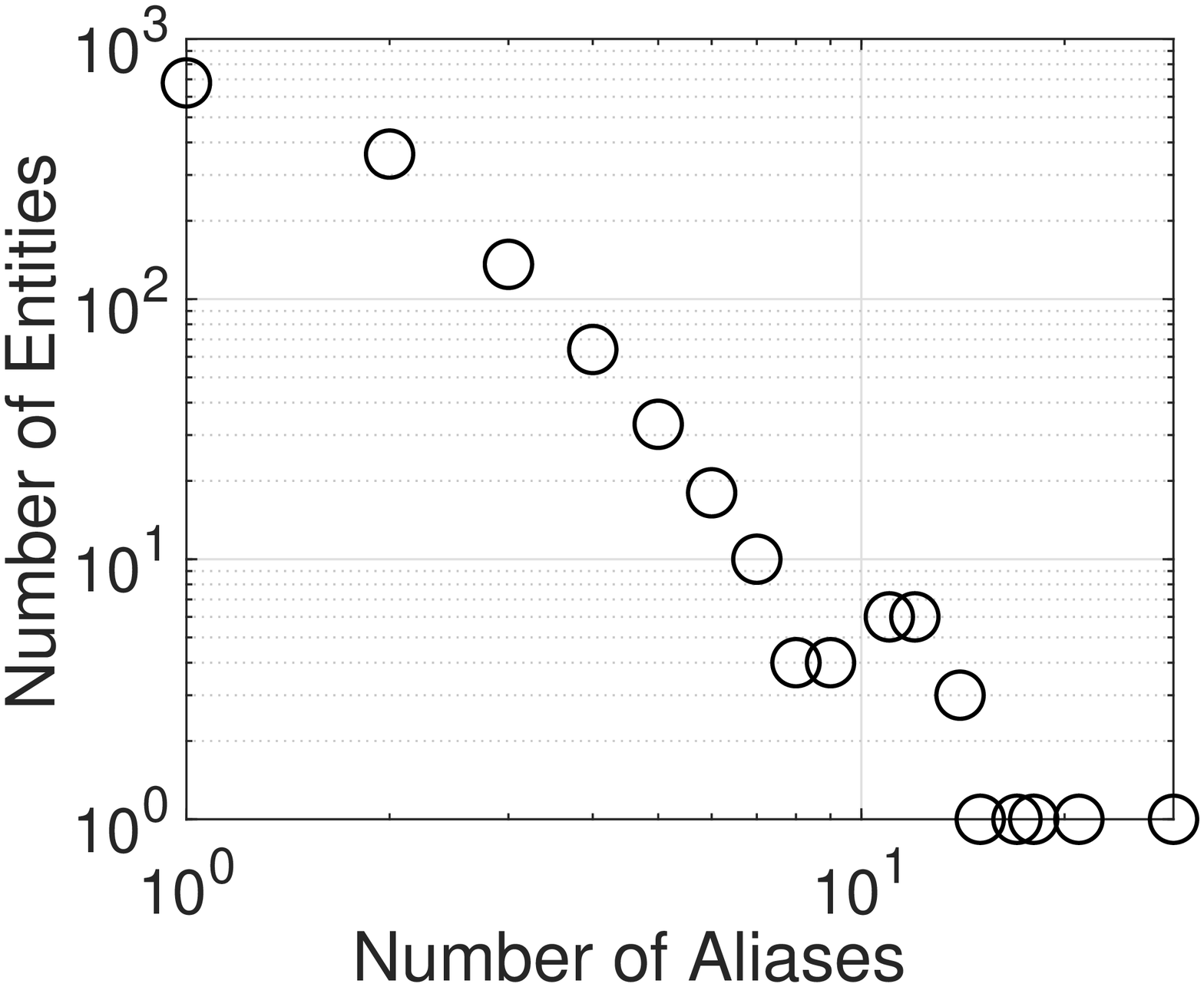} }  
	\subfigure[GDA] 	{\includegraphics[width=0.32\textwidth, draft=false]{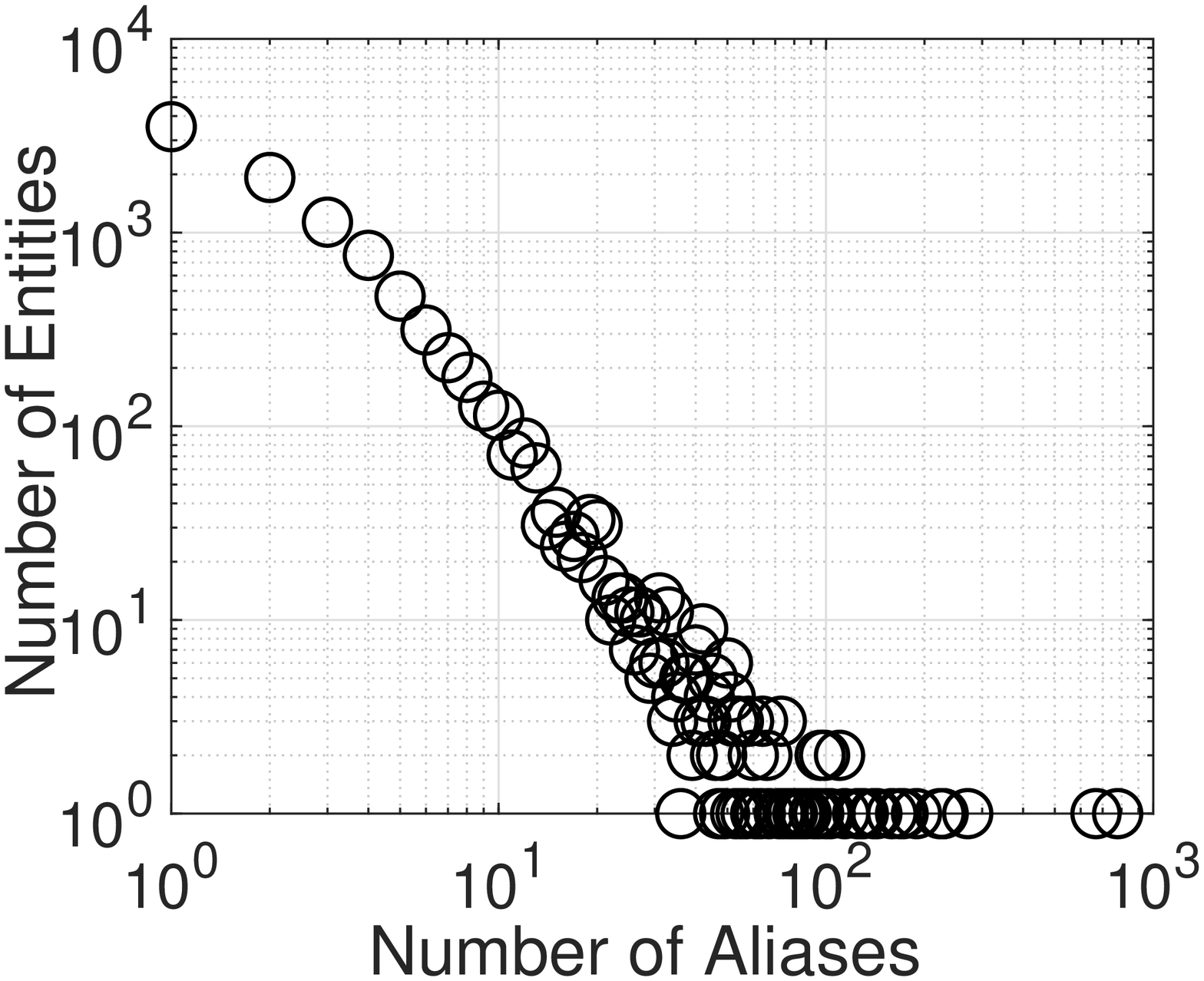}} 
	\caption{Statistics of entity aliases on three datasets.}
	\label{fig:aliases}
\end{figure*}

Figure~\ref{fig:aliases} plots the distribution of number of entities to number of aliases on three commonly-used datasets. 
For DocRED, we can observe that most of entities have only one alias and 4,745 entities have more than one alias. 
The maximum number of aliases is 10. 
For CDR, 650 entities (account for 48.95\%)  have more than one alias and the maximum number of aliases is 29. 
For GDA, 5,927 entities (account for 62.73\%)  have more than one alias and the maximum number of aliases is 778. 
CDR and GDA have more diverse aliases than DocRED, because DocRED is constructed from Wikipedia while CDR and GDA are constructed from biomedical text.   
Linking diverse aliases of an entity to its identifier is a challenging task in a long document. 
Our findings identify the strong (almost untenable) assumption of existing DocRE models that all the aliases (\ie unique mentions) of an entity are successfully normalized (\ie linked to its unique identifier).

\paratitle{\texttt{\textbf{(RQ3)}}}: \textbf{Do the aliases of an entity vary widely in commonly-used DocRE datasets?}

\texttt{\textbf{RQ2}} already showed that an entity may have multiple aliases. 
For example, an entity in GDA has $778$ unique aliases. 
In this Section, we investigate whether the aliases of an entity vary widely. 
Table~\ref{tab:topaliases} shows details of entity aliases ranked by numbers of aliases in the three datasets. 
For DocRED, the variation of aliases is slight because the genre of text is from formal articles.
Although DocRED is manually annotated by human beings, there are still some annotation errors on entity linking. 
As shown in Table~\ref{tab:topaliases}, the entity (Q180611, Azpeitia) is linked to many wrong aliases such as ``United States'' and ``Chile''. 
This observation confirms that document-level entity linking is a very challenging task. 
For biomedical datasets, chemical entities have a slight variation of aliases while gene and disease entities have a huge variation of aliases. 
In addition, there are many abbreviations for biomedical entities.  
Thus, effective NER and entity linking are key preconditions in modeling DocRE. 
We will further investigate the model usability with NER and entity linking in Section \S~\ref{sec:usability}.

\begin{table*}[t]
	\centering
	\small
	\begin{tabular}{c|p{1.5cm}|c|p{10.8cm}}
		\toprule
		\textbf{Rank} & \textbf{IDs} & \textbf{\#Aliases}& \textbf{Details of Entity Aliases}                                                                     \\ \midrule
		
		\multicolumn{4}{c}{\textbf{DocRED Dataset}}   \\ \hline

		\multirow{2}{*}{1}    & Q180611 (LOC) &  \multirow{2}{*}{10} & Azpeitia, Guipuzcoa, Cuba, Mexico, Azkoitia, Basque Country, United States, Argentina, Chile, Spain\\ \hline
		
		\multirow{2}{*}{2}    & Q544565 (LOC) &  \multirow{2}{*}{10} & Qu, Yuxi River, Jiuxi River, Zhuji River, Ni River, Eshan River, Liucun River, Huaxi River, Qu River, Zhou River\\ \hline

		\multirow{1}{*}{3}    & Q3738980 (LOC) &  \multirow{1}{*}{8} &Toding, Tsanda, Tsada, Tholing, Zanda, Toling, Zada, Tuolin\\ \hline
		

		\multirow{2}{*}{4}    & Q12274473 (MISC) &  \multirow{2}{*}{8} &wazīrwāla, Waziri, Maseedwola, Wazirwola, Dawarwola, Wazir, of the Wazirs, Waziri Pashto \\ \hline

		\multicolumn{4}{c}{\textbf{CDR Dataset}}   \\ \hline
		
		\multirow{2}{*}{1}    & D016572 (Chemical) &  \multirow{2}{*}{7} &cyclosporine, cyclosporin, CsA, Cyclosporine, CyA, cyclosporin A, cyclosporine A  \\ \hline
		
		\multirow{2}{*}{2}    & D014635 (Chemical) &  \multirow{2}{*}{7} &divalproex sodium, VPA, sodium valproate, Valproic acid, Valproate, valproic acid, valproate \\ \hline
		
		\multirow{3}{*}{1}    & D007674 (Disease) &  \multirow{3}{*}{29} &renal damage, CAN, nephrotoxic, renal dysfunctio, renal injury, Nephrotoxicity, kidney diseases, liver or kidney disease, cardiac and renal lesions, glomerular injury, kidney damage, ...omit... \\ \hline
		
		\multirow{3}{*}{2}    &D056486 (Disease) &  \multirow{3}{*}{21} &Hepatitis, drug-induced hepatitis, acute hepatitis-like illness, liver damage, hepatotoxicity,  cholestatic hepatitis, hepatic damage,  hepatocellular injury, Toxic hepatitis, Granulomatous hepatitis, ...omit... \\ \hline

		\multicolumn{4}{c}{\textbf{GDA Dataset}}   \\ \hline
		
		\multirow{3}{*}{1}    & 348 \,\,  (Gene) &  \multirow{3}{*}{114} &apolipoprotein e4, APOE*4, ApoE2, apolipoprotein gene E4 allele, ApoE-4, apoE 4, apolipoprotein-E gene, Apolipoprotein E-epsilon4, factor--apolipoprotein E, Apolipoprotein (apo)E, ...omit... \\ \hline

		\multirow{3}{*}{2}    & 7124 (Gene) &  \multirow{3}{*}{83} &tumor necrosis factor alpha, tumor necrosis factor beta, Interleukin-1 and tumor necrosis factor-alpha, tumor necrosis factor alpha, TNF-)a, Tumor Necrosis Factor, TNF-308G/A, miR-21, IL6 and TNF, ...omit... \\ \hline

		\multirow{4}{*}{1}    & D030342 (Disease) &  \multirow{4}{*}{778} &inherited defect of fatty acid oxidation, genetic haemochromatosis, inherited skin disorders, A-related disorders, autosomal-recessive pleiotropic disorder, autosomal dominant juvenile ALS, ...omit... \\ \hline
		
		\multirow{4}{*}{2}    &D009369 (Disease) &  \multirow{4}{*}{668} &mammary tumors, tumor suppressor genes, MSI-H cancers, rectal cancers, predominant in lung tumour, early-stage prostate cancer, Tumour-necrosis, Cervical cancer, Malignant tumors, distal tumors, ...omit... \\ 
		\bottomrule                                                             
	\end{tabular}
	\caption{Details of entity aliases ranked by number of aliases.}
	\label{tab:topaliases}
\end{table*}

\section{Check on Model Robustness}
\label{sec:robustness}

Most of existing DocRE models are proposed based on the strong assumptions of mention annotations as shown in Section \S~\ref{sec:dataannotation}. 
In this Section, we are interested in the following research question: 

\paratitle{\texttt{\textbf{(RQ4)}}}: \textbf{Are neural DocRE models robust to entity mention attacks?}

To answer \texttt{\textbf{RQ4}}, we adopt  \textbf{behavioral probing}~\cite{DBLP:conf/acl/LasriPLPC22, DBLP:conf/naacl/0003HNC22} to observe a model's behaviors by studying the model's predictions on attacking datasets. 
That is, attacks are only added at test time and are not available during model training.

\subsection{Attacking Target Models}

We investigate three typical DocRE models: 
(1) \textbf{BiLSTM-Sum}~\cite{DBLP:conf/acl/YaoYLHLLLHZS19} which uses BiLSTM to encode the document and computes the representation of an entity by summing the representations of all mentions.
(2) \textbf{GAIN-Glove}~\cite{DBLP:conf/emnlp/ZengXCL20} which constructs a heterogeneous mention-level graph and an entity-level graph to capture document-aware features and uses GloVe~\cite{DBLP:conf/emnlp/PenningtonSM14} as word embeddings. 
(3) \textbf{BERT-Marker}~\cite{DBLP:conf/aaai/Zhou0M021,DBLP:journals/corr/abs-2102-01373} which takes BERT as the encoder and inserts special entity symbols before and after entities. 
More details of attacking target models can be found in Appx. \S~\ref{app:baselines}. 

\begin{table*}[t]
	\centering
	\scalebox{0.97}{
	\begin{tabular}{ll|rr|rr|rr}
		\toprule
		\multirow{2}{*}{\textbf{Model}}  & \multirow{2}{*}{\textbf{Attack}} & \multicolumn{2}{c|}{\textbf{DocRED}}                  & \multicolumn{2}{c|}{\textbf{CDR}}                     & \multicolumn{2}{c}{\textbf{GDA}}                     \\ \cline{3-8}
		&                & \textbf{F1}\%                 & $\Delta$\%           & \textbf{F1}\%                 & $\Delta$\%           & \textbf{F1}\%                & $\Delta$\%           \\  \midrule
		\multirow{6}{*}{BiLSTM-Sum} & No Attack                       & 49.32               &  -                &     53.67         &    -                 & 75.87                &       -         \\ 
		
		& \texttt{DrpAtt}                       &    42.55               & -13.73             &       48.34             & -9.93               &  66.55                   &-12.28        \\
		&  \texttt{BryAtt}&   39.04	&-20.84	&39.23	&-26.91	&57.30	&-24.48      \\
		&  \texttt{CorAtt}&   37.21	&-24.55	&28.86	&-46.23	&31.32&	-58.72   \\ 
		& \texttt{MixAtt}                       &     32.67	& -33.76	& 22.26	& -58.52	& 24.88	& -67.21      \\  \hline

		\multirow{6}{*}{GAIN-Glove} & No Attack                       & 54.91             &  -                &     55.13         &    -                 & 78.65                &       -         \\ 
		
		& \texttt{DrpAtt}                       &   48.17 &	-12.27&	50.76	&-7.93&	59.61&	-24.21   \\
		&  \texttt{BryAtt}&  41.82	&-23.84&	36.33	&-34.10&	45.92&	-41.61     \\
		&  \texttt{CorAtt}&    32.34&	-41.11&	27.40&	-50.30&	33.24&	-57.74  \\ 
		& \texttt{MixAtt}          & 28.56	&-47.99	&18.34	&-66.73&	23.52&	-70.10  \\  \hline

		\multirow{6}{*}{BERT-Marker}  & No Attack                       & 59.82              &  -               &     64.47         &    -                 & 82.71               &       -         \\ 
		
		& \texttt{DrpAtt}                       &   47.34	&-20.86	&25.57	&-60.34	&36.43	&-55.95        \\
		&  \texttt{BryAtt}&  41.45	&-30.71	&21.46	&-66.71	&24.55	&-70.32     \\
		&  \texttt{CorAtt}&  25.86	&-56.77	&16.93	&-73.74	&19.04	&-76.98 \\ 
		& \texttt{MixAtt}       & 18.34	&-69.34	&9.34&	-85.51&	13.55&	-83.62           \\

		\bottomrule
		
	\end{tabular}}
\caption{Results of mention attacks on three datasets.  $\Delta$\% indicates the relative performance changes between mention attacks and the original input (``No Attack'').}
\label{tab:attackresults}
\end{table*}

\subsection{Attack Construction}
In this work, we focus on entity mention attacks which add data perturbations by taking into account different types of wrongly-detected mentions. 
The ultimate goal is to test the model robustness under different mention attacks. 
Therefore, we construct four types of attacks:
(1) \texttt{DrpAtt}: we simply drop 50\% of mentions of an entity if the entity has more than one mention. This attack is designed to simulate the case of missed detections in NER systems. 
(2) \texttt{BryAtt}: we slightly move the ground boundaries of 50\% of mentions of an entity if the entity has more than one mention (\eg ``$\llbracket$Spanish Civil War$\rrbracket_\text{MISC}$ in'' is changed to  ``Spanish  $\llbracket$Civil War$\rrbracket_\text{MISC}$ in'').   
(3) \texttt{CorAtt}: we intentionally make the coreference (\ie entity linking) of an entity wrong (\ie $50\%$ of mentions of an entity are wrongly coreferential if the entity has more than one mention). 
(4) \texttt{MixAtt}: this attack is the mix of aforementioned three attacks. More attack details can be found in Appx. \S~\ref{app:attacks}.

\subsection{Attacking Results and Analysis}

Table~\ref{tab:attackresults} reports the performance on various entity mention attacks for three attacking target models. 
We have the following observations: 

First, all target models are significantly affected by the four attacks, with relative F1 drops from 7.93\% to 85.51\%. 
Overall, GAIN-Glove and BERT-Marker are more vulnerable than BiLSTM-Sum. 
This is because BERT-Marker requires accurate mention positions for inserting entity markers and  GAIN-Glove needs the information of mention positions and normalization for constructing heterogeneous graphs. 
More specifically, BERT-Marker averagely suffers drops of 44.42\%, 71.58\%, and 71.72\% across all attacks on DocRED, CDR and GDA, respectively. 
BiLSTM-Sum averagely suffers drops of 23.22\%, 35.40\%, and 40.67\% across all attacks on DocRED, CDR and GDA, respectively.

Second, the \texttt{MixAtt} attack leads to more significant drops in performance for all attacking target models. 
\texttt{CorAtt} is more significant to impact robustness than  \texttt{BryAtt} and \texttt{DrpAtt}. 
For instance, \texttt{CorAtt} leads to relative drops of 40.81\%,  56.76\% and 64.48\% across three target models on DocRED, CDR and GDA, respectively. 
\texttt{DrpAtt} leads to relative drops of 15.62\%,  26.07\% and 30.81\% across three target models on DocRED, CDR and GDA, respectively. 
Our empirical results clearly show that the information of entity coreference, boundary and position plays an important role in DocRE.

Overall, based on the robustness evaluation in Table~\ref{tab:attackresults},  we can answer \texttt{\textbf{RQ4}}: Most of neural DocRE models are far away from robustness to entity mention attacks. 
Therefore, it has some realistic significance to challenge current problem setups regarding data annotation assumptions in DocRE and to improve the robustness of DocRE models on entity mention attacks.

\section{Check on Model Usability}
\label{sec:usability}
In this Section, we investigate this realistic situation:  DocRE models are already trained and training data is unavailable. We want to extract same relations on unseen raw text using these models. 
The goal is to deploy the already-trained DocRE models in other NLP applications. 
Here, we are interested in the following research question: 

\paratitle{\texttt{\textbf{(RQ5)}}}: \textbf{Are existing DocRE models easily adopted in real-world DocRE scenarios?}

To answer \texttt{\textbf{RQ5}}, a necessary step is that whether we can process the raw text with the format as DocRE models trained on. 
This preprocessing procedure involves two crucial systems: Named Entity Recognition (NER) and Entity Linking.

\subsection{Check on NER}
\label{subsec:ner}
\paratitle{Setups.}
Assume that DocRE models are already trained and the training sets are unavailable. 
We take the raw text of development set of DocRED, and test sets of CDR and GDA as the unseen data. 
We use strict match metrics (\ie entity boundary and type are both correctly detected) to measure agreement between the annotations we preprocessed 
and existing ground truth annotations. 

\paratitle{NER Systems.}
For DocRED, we adopt three off-the-shelf NER systems: Flair~\cite{akbik2019flair} and spaCy\footnote{\url{https://spacy.io/}} and Stanza~\cite{DBLP:conf/acl/QiZZBM20}. 
For CDR and GDA, we adopt three biomedical NER systems: HunFlair~\cite{DBLP:journals/bioinformatics/WeberSMHLA21}, Stanza biomedical models~\cite{DBLP:journals/corr/abs-2007-14640} and Scispacy~\cite{DBLP:conf/bionlp/NeumannKBA19}. 
More details of NER systems can be found in Appx. \S~\ref{app:ner}. 

\paratitle{Results on NER.}
Table~\ref{tab:ner} reports experimental results of NER systems on the three datasets. 
For DocRED, Flair achieves the best performance by the F1 score of 63.47\%.   
Although the genre of DocRED is the formal text (\ie Wikipedia), the state-of-the-art NER systems are still unable to achieve decent performance on DocRED. 
HunFlair gets the best performance on the biomedical datasets because it trained on harmonized versions of 31 biomedical datasets. 
\begin{table}[t]
	\centering
	\scalebox{0.95}{
		\begin{tabular}{c|c|ccc}
			\toprule
			
			\multirow{2}{*}{\textbf{Dataset}} &\multirow{2}{*}{\begin{tabular}[c]{@{}c@{}}\textbf{NER}\\ \textbf{System}\end{tabular}} & \multicolumn{3}{c}{\textbf{Strict Match} (\%)} \\
			&                        &\textbf{ P}    &\textbf{R}      & \textbf{F1}      \\
			
			\midrule
			\multirow{3}{*}{DocRED} 
			& Flair          &  62.88& 64.07  & 63.47   \\
			& spaCy          & 62.86  & 59.58  &  61.17  \\
			& Stanza          & 56.96  &58.44   &57.69    \\  \hline
			\multirow{3}{*}{CDR}                          & HunFlair           &94.59  & 94.14  &   94.36 \\ 
			& Stanza           & 86.80 & 87.94  &87.37  \\ 
			& ScispaCy           &84.93   &80.32   &82.56    \\  \hline
			\multirow{3}{*}{GDA}                              &  HunFlair          & 79.11& 84.74 &81.83   \\
			&  Stanza          & 69.87& 79.70 &74.47   \\
			&  ScispaCy          & 68.61& 64.61 &66.55   \\ \bottomrule
	\end{tabular}}
	\caption{Results of NER systems.}
	\label{tab:ner}
\end{table}

\subsection{Check on Entity Linking}
\label{subsec:linking}

\paratitle{Setups.}
We examine the capability of entity linking systems on reproducing ground truth annotations for development/test sets of DocRED, CDR and GDA.
We choose the strict match as the metric that a linking prediction is regarded as correct only if all mentions of an entity are correctly linked to the entity. 

\paratitle{Entity Linking Systems.}
Unlike NER systems, there are very few off-the-shelf linking systems available. 
We choose TagMe~\cite{DBLP:conf/cikm/FerraginaS10} as the linker for DocRED, and Scispacy~\cite{DBLP:conf/bionlp/NeumannKBA19} for CDR and GDA. 
More details of entity linking systems can be found in Appx. \S~\ref{app:linking}.

\paratitle{Results on Entity Linking.}
Table~\ref{tab:linking} reports experimental results of entity linking systems on the three datasets. 
For TagMe, the precision increases gradually with the increase of the value of $\rho$ (confidence score), while the recall decreases as $\rho$ increases. 
The best F1 on DocRED is only 38.7\% with a confidence score of 0.3. 
Scispacy achieves F1 scores of 58.1\% and 34.3\% using umls for CDR and GDA, respectively.
One key observation drawn from Table~\ref{tab:linking} is that document-level entity linking is a challenging task and existing linking systems commonly perform poorly on this task.   
\begin{table}[t]
	\centering
	\scalebox{0.95}{
		\begin{tabular}{c|c|ccc}
			\toprule
			
			\multirow{2}{*}{\textbf{Dataset}} &\multirow{2}{*}{\begin{tabular}[c]{@{}c@{}}\textbf{Linking}\\ \textbf{System}\end{tabular}} & \multicolumn{3}{c}{\textbf{Strict Match} (\%)} \\
			&                        &\textbf{ P}    &\textbf{R}      & \textbf{F1}      \\

			\midrule
			\multirow{4}{*}{DocRED} & TagMe, $\rho$=0.1      & 24.2 & 42.5 &30.8    \\
			& TagMe, $\rho$=0.2          & 35.0  &   38.6&   36.7 \\
			& TagMe, $\rho$=0.3         & 45.7  &   33.5&   38.7 \\
			& TagMe, $\rho$=0.4        &  52.4 &27.8   &   36.4 \\
			& TagMe, $\rho$=0.5        &49.7  & 12.4  & 19.8 \\  \hline
			\multirow{2}{*}{CDR} 
			&    ScispaCy, mesh        &42.4   & 60.6  & 49.9   \\                   
			&    ScispaCy, umls        &53.7   & 63.3  & 58.1   \\ 
			\hline
			\multirow{2}{*}{GDA}                      &   ScispaCy, mesh       &  31.5 &28.4   & 29.8   \\ 
			&   ScispaCy, umls       &  30.9 &38.6   & 34.3   \\ 
			\bottomrule
	\end{tabular}}
	\caption{Results of entity linking systems. $\rho$ is the confidence score (annotations that are below the threshold will be discarded). ``mesh'' and ``umls'' mean that entities are linked to the Medical Subject Headings and the Unified Medical Language System, respectively.}
	\label{tab:linking}
\end{table}

Based on empirical results of Sections \ref{subsec:ner} and \ref{subsec:linking}, we can answer \texttt{\textbf{RQ5}}: Most of existing DocRE models are difficult to be adopted in real-world DocRE scenarios due to the need of input preparation for each pipeline module  and the accumulation of errors in NER and entity linking systems.

\section{Discussion}
\label{sec:discussions}

\paratitle{Let's Stop Simplifying Problem Setups.}
As summarized in Table~\ref{tab:relatedsurvey}, recent advances from the past four years have claimed significant progress in DocRE performance.
However, our study shows that the actual improvements are attributable to a strong or even untenable assumption where all entities are perfectly typed, localized and normalized. 
Therefore, high F1 scores on leaderboards do not mean that the task of DocRE has been solved. 
Based on our findings~(\S \ref{sec:robustness} and \S \ref{sec:usability}), the simplified problem setups cannot cover realistic scenarios. 
Even worse, the problem simplification significantly hurts the usability of deploying DocRE models in real-world end-user NLP applications. 
We call attentions on the community to address the real DocRE problem under the open-world assumption, rather than to push up the boundaries of simplified benchmarks for leading leaderboards.

\paratitle{Let's Model DocRE in the Wild.}
As shown in Section \S~\ref{sec:usability}, it is very difficult to produce accurate data formats as existing DocRE models trained on.  
Thus, given a new document, we are still unable to easily deploy existing trained DocRE models to extract same types of relations, let alone unseen relations.  
Recently, some studies~\cite{DBLP:conf/emnlp/CabotN21,DBLP:conf/eacl/EbertsU21,john2022} have started exploring the direction of jointly extracting entities and relations at document level. 
However, the end-to-end performance at document level is much worse than the performance at sentence level. 
Our empirical findings call more attentions on developing high-performance end-to-end DocRE models and  more attentions on modeling DocRE in the wild, rather than in an unrealistic Utopian world.

\section{Conclusion}
In this paper, we try to answer whether the performance gains recent DocRE models claimed are actually true. 
We took a comprehensive literature review of DocRE models and a thorough examination of popular DocRE datasets. 
We investigated the model robustness under four types of mention attacks and the model usability under a more realistic setting. 
Our findings call future efforts on modeling DocRE in the wild. 

\section*{Limitations}
We have discussed the implications of our research in Section~\ref{sec:discussions}.
In this Section, we further discuss the threats to validity of our study.

 \begin{itemize} [noitemsep,nolistsep]
 	\item \textbf{Threats to Internal Validity}:
 	The main internal threat to the validity of our research comes from \texttt{\textbf{(RQ3)}} where we present a qualitative study on the variation of aliases. 
 	We are unable to cover all cases in the qualitative study. For example, the entity of D030342 (Disease) in Table~\ref{tab:topaliases} has $778$ unique aliases. It is impossible to show all aliases to readers.  
 	To help mitigate this threat, we try to show as many examples as possible in a limited space.  
 	\item \textbf{Threats to External Validity}: The main threat to external validity arises from the potential bias in the selection of experimental datasets, attacking target models and off-the-shelf NER and Entity Linking tools. To mitigate this threat, we experiment with multiple datasets, models and tools. For experimental datasets, we choose the three most popular DocRE datasets (\ie DocRED, CDR, and GDA). We believe that these three datasets are broadly representative in this research community. 
 	For attacking target models, we choose three typical models ranging from non-contextualized sequence-based to graph-based, and to contextualized Transformers models. 
 	For off-the-shelf NER/Linking tools, we comprehensively investigate five state-of-the-art NER taggers and two entity linkers.   
 \end{itemize}

\section*{Ethical Considerations}
As our goal of this study is to challenge current problem setups of DocRE, we heavily rely upon existing well-known datasets, models and NLP tools. 
We only claim that our findings may hold on similar datasets or domains. 
We acknowledge the risk of generalizability of our findings on other privacy-sensitive datasets or specific domains. 
In general, we suggest that practitioners repeat all experiments following our procedures when using other corpora. 

\bibliographystyle{acl_natbib}
\bibliography{bibfile}

\appendix

\section*{Appendix}

\section{Additional Examples of Data Annotations}
\label{app:moreillustrating}

\begin{figure*}[t]		
	\centering
	\includegraphics[width=0.95\textwidth,draft=false]{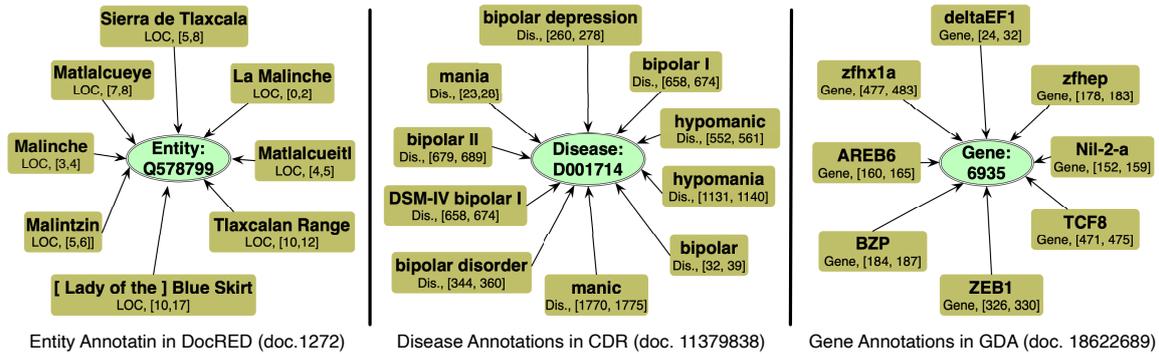}	
	\caption{Additional examples of data assumption in three popular DocRE datasets. Entities are annotated with types (LOC/Disease/Gene), positions ([start, end]) and unique identifiers (Q578799/D001714/6935).  }
	\label{fig:3datasets}
\end{figure*}
Our study identifies a strong or even untenable assumption in DocRE. 
To give more intuitive sense, Figure~\ref{fig:3datasets} shows additional examples of data assumption in three popular DocRE datasets.
Specifically, the eight entity mentions (\ie \textit{Tlaxcalan Range, Matlalcueitl, [ Lady of the ] Blue Skirt, Malintzin, Sierra de Tlaxcala, Malinche, La Malinche, Matlalcueye}) are annotated with types and positions, then linked to a unique identifier in the DocRED corpus. 
The ten entity mentions (\textit{mania, bipolar II, bipolar I, bipolar depression, hypomanic, hypomania, DSM-IV bipolar I, bipolar, manic, bipolar disorder}) are typed, localized and normalized in the CDR corpus. 
The eight entity mentions (\textit{deltaEF1, zfhx1a, zfhep, AREB6, Nil-2-a, BZP, TCF8, ZEB1}) are typed, localized and normalized in the GDA corpus. 
Most of existing DocRE models are developed based on the assumption that all entity mentions are perfectly typed, localized and normalized.

\section{More Experimental Details}

\subsection{Attacking Target Models}
\label{app:baselines}

\paratitle{BiLSTM-Sum.}
BiLSTM-Sum~\cite{DBLP:conf/acl/YaoYLHLLLHZS19} uses a bidirectional LSTM to encode documents and computes the representation of an entity by summing the representations of all mentions.
The embeddings from \texttt{glove.840B.300d}\footnote{\url{https://nlp.stanford.edu/projects/glove/}} are used to initialize model vocabularies for DocRED, CDR and GDA. 
All word embeddings and model parameters are learnable during training.   
Hyperparameters are tuned on the development set for each dataset respectively.

\paratitle{GAIN-Glove.}
GAIN-Glove~\cite{DBLP:conf/emnlp/ZengXCL20} constructs a heterogeneous mention-level graph to model complex interaction among different mentions across the document.
Then a path reasoning mechanism is proposed to infer relations between entities based on another constructed entity-level graph.
We implement GAIN-Glove with 2 layers of GCN and the dropout rate of 0.6 based on the codes\footnote{\url{https://github.com/DreamInvoker/GAIN}}.
The embeddings from \texttt{glove.840B.300d}\footnote{\url{https://nlp.stanford.edu/projects/glove/}} are used  for DocRED, CDR and GDA.

\paratitle{BERT-Marker.}
BERT-Marker~\cite{DBLP:conf/aaai/Zhou0M021,DBLP:conf/naacl/ZhongC21,DBLP:journals/corr/abs-2102-01373} first inserts special entity symbols (\ie \texttt{[ent]} and \texttt{[/ent]}) before and after entities, then encodes the whole document using the pretrained BERT. 
The representation of token \texttt{[CLS]} is used for classification. 
In particular, we use the checkpoint \texttt{bert-base-uncased}\footnote{\url{https://huggingface.co/bert-base-uncased}} for DocRED, and the checkpoint \texttt{allenai/scibert\_scivocab\_uncased}\footnote{\url{https://huggingface.co/allenai/scibert_scivocab_uncased}} for CDR and GDA.

All attacking target models are implemented with \texttt{PyTorch}\footnote{\url{https://pytorch.org/}} and \texttt{Accelerate}\footnote{\url{https://github.com/huggingface/accelerate}}, and  trained on one DGX machine, totally equipped with 80 Intel(R) Xeon(R) CPU E5-2698 v4 @ 2.20GHz processor cores, 400 GB of RAM, and 8 NVIDIA Tesla V100-32GB GPUs.

\subsection{Attack Details}
\label{app:attacks}

In total, we construct four types of attacks, \ie, \texttt{DrpAtt}, \texttt{BryAtt}, \texttt{CorAtt} and \texttt{MixAtt}, to check the robustness of attacking target models.

\paratitle{\texttt{DrpAtt}.}
Missing some entities is a very common phenomenon for most of NER systems.
\texttt{DrpAtt} is constructed to investigate the effect of missed mentions. 
If an entity has more than one mention, we simply drop $50\%$ of mentions of the entity. 

\paratitle{\texttt{BryAtt}.}
Some entities are complex and nested in natural language. 
Detecting boundaries precisely is not a trivial task. 
\texttt{BryAtt} is constructed to investigate the effect of wrongly-detected entity boundaries.  
If an entity has more than one mention, we slightly move the ground boundaries of $50\%$ of mentions of the entity.

\paratitle{\texttt{CorAtt}.}
The document-level coreference resolution is a challenging task in DocRE. 
Most of existing DocRE models are developed on benchmark datasets where entity coreference is manually annotated. 
\texttt{BryAtt} is constructed to investigate the effect of wrongly-coreferential mentions.  
We intentionally make the coreference results (\ie entity linking) of an entity wrong (\ie $50\%$ of mentions of an entity are wrongly coreferential if the entity has more than one mention).

\paratitle{\texttt{MixAtt}.}
This type of attack is the mix of aforementioned three attacks.

\subsection{NER Systems}
\label{app:ner}
In Section~\ref{subsec:ner}, we adopt five off-the-shelf NER systems in our experiments. 

\paratitle{Flair.}
Flair\footnote{\url{https://github.com/flairNLP/flair}} is a very simple framework for state-of-the-art NLP and developed by Humboldt University of Berlin and friends.
We use the \texttt{ner-english-ontonotes-large}\footnote{\url{https://huggingface.co/flair/ner-english-ontonotes-large}} model for DocRED.

\paratitle{spaCy.}
spaCy\footnote{\url{https://spacy.io/}} is a library for advanced Natural Language Processing in Python and Cython.
We use the \texttt{en\_core\_web\_trf}\footnote{\url{https://spacy.io/models/en\#en_core_web_trf}} model for DocRED. 

\paratitle{Stanza.}
Stanza\footnote{\url{https://stanfordnlp.github.io/stanza/}} is a collection of accurate and efficient tools for the linguistic analysis of many human languages, developed by Stanford NLP Group. 
General domain, biomedical $\&$ clinical models are available in Stanza. 
We use the \texttt{ontonotes}\footnote{\url{https://stanfordnlp.github.io/stanza/ner_models.html}} for DocRED,   \texttt{bc5cdr}\footnote{\url{https://stanfordnlp.github.io/stanza/available_biomed_models.html}} for CDR, 
\texttt{bc5cdr} and \texttt{bionlp13cg} for GDA.

\paratitle{HunFlair.}
HunFlair\footnote{\url{https://github.com/flairNLP/flair/blob/master/resources/docs/HUNFLAIR.md}} is a state-of-the-art NER tagger for biomedical texts. 
It contains harmonized versions of 31 biomedical NER datasets.
We use \texttt{hunflair-chemical} and \texttt{hunflair-disease} for CDR, 
\texttt{hunflair-gene} and \texttt{hunflair-disease} for GDA.\footnote{\url{https://github.com/flairNLP/flair/blob/master/flair/models/sequence_tagger_model.py\#L751}} 

\paratitle{ScispaCy.}
ScispaCy\footnote{\url{https://allenai.github.io/scispacy/}} is a Python package containing spaCy models for processing biomedical, scientific or clinical text.
We use \texttt{en\_ner\_bc5cdr\_md} for CDR. 
We use \texttt{en\_ner\_bc5cdr\_md}, and \texttt{\texttt{en\_ner\_bionlp13cg\_md}} for GDA.\footnote{\url{https://github.com/allenai/scispacy}}

\subsection{Entity Linking Systems}
\label{app:linking}
Comparing with flourishing NER systems, there are very few entity linking systems available.  
We adopt two widely-used entity linking systems in our experiments. 

\paratitle{TagMe.}
TagMe\footnote{\url{https://sobigdata.d4science.org/web/tagme/tagme-help}} is a powerful tool that identifies on-the-fly meaningful substrings (called ``spots'') in an unstructured text and link each of them to a pertinent Wikipedia page in an efficient and effective way. 
We use the official Python TagMe API wrapper\footnote{\url{https://github.com/marcocor/tagme-python}} for DocRED.
The confidence scores (annotations that are below the threshold will be discarded) are experimented among $[0.1,0.2,0.3,0.4,0.5]$.

\paratitle{Entity Linker in ScispaCy.}
Entity Linker in ScispaCy\footnote{\url{https://github.com/allenai/scispacy\#entitylinker}} is a spaCy component which performs linking to a knowledge base. The linker simply performs a string overlap - based search (char-3grams) on named entities, comparing them with the concepts in a knowledge base using an approximate nearest neighbours search.
For CDR and GDA datasets, we explore the following two knowledge bases: 
\begin{itemize} [noitemsep,nolistsep]
\item umls: Links to the Unified Medical Language System, levels 0,1,2 and 9. This has 3 million concepts.
\item mesh: Links to the Medical Subject Headings. This contains a smaller set of higher quality entities, which are used for indexing in Pubmed. MeSH contains ~30k entities. 
\end{itemize}

\section{License}
DocRED is released under The MIT license.
GDA  is released uner The GNU Affero General Public License. 
GAIN  is released under The MIT License. 
Flair  is released under The MIT License. 
spaCy  is released under The MIT License. 
Stanza is Licensed under The Apache License 2.0.
HunFlair is Licensed under The MIT License. 
ScispaCy is Licensed under The Apache License 2.0. 
TagMe is Licensed under The Apache License 2.0. 
PyTorch is with The Copyright (c) 2016 -  Facebook, Inc (Adam Paszke).
Huggingface Transformer models are released under The Apache License 2.0.
All the scientific artifacts are consistent with their intended uses.

\end{document}